\documentclass[runningheads]{llncs}
\pdfoutput=1 

\newcommand{\model}{\textsf{FedX}}

\usepackage{graphicx}
\usepackage{multirow}
\usepackage{comment}
\usepackage{amsmath,amssymb} 
\usepackage{color}
\usepackage{caption}
\usepackage{subcaption}
\usepackage{mathtools}
\usepackage[algo2e,ruled]{algorithm2e}
\usepackage{algorithm}
\usepackage{booktabs}
\usepackage{subcaption}
\usepackage{comment}
\usepackage{xcolor}
\usepackage{hyperref}
\hypersetup{
    colorlinks=true,
    linkcolor=black,
    filecolor=magenta,      
    urlcolor=black,
}
\usepackage[accsupp]{axessibility}  

\DeclareMathOperator*{\argmin}{arg\,min}
\DeclarePairedDelimiter\norm{\lVert}{\rVert}

\begin{document}
\pagestyle{headings}
\mainmatter
\def\ECCVSubNumber{3932}  

\title{FedX: Unsupervised Federated Learning \\ with Cross Knowledge Distillation}  

\titlerunning{FedX: Unsupervised Federated Learning
with Cross Knowledge Distillation}
%
\author{Sungwon Han\thanks{Equal contribution to this work.}\inst{1,2}\orcidID{0000-0002-1129-760X} \and
Sungwon Park$^\star$\inst{1,2}\orcidID{0000-0002-6369-8130} \and
Fangzhao Wu\inst{3}\orcidID{0000-0001-9138-1272} \and
Sundong Kim\inst{2}\orcidID{0000-0001-9687-2409} \and
Chuhan Wu\inst{4}\orcidID{0000-0001-5730-8792} \and
Xing Xie\inst{3}\orcidID{0000-0002-8608-8482} \and
Meeyoung Cha\inst{2,1}\orcidID{0000-0003-4085-9648}
}
\authorrunning{S. Han et al.}
%
\institute{School of Computing, KAIST \\
\email{\{lion4151, psw0416\}@kaist.ac.kr}\\ \and
Data Science Group, Institute for Basic Science \\
\email{\{sundong, mcha\}@ibs.re.kr} \\ \and
Microsoft Research Asia \\
\email{wufangzhao@gmail.com, xingx@microsoft.com} \\ \and
Tsinghua University \\
\email{wuchuhan15@gmail.com}}
\maketitle
\begin{abstract}
This paper presents \model{}, an unsupervised federated learning framework. Our model learns unbiased representation from decentralized and heterogeneous local data. It employs a two-sided knowledge distillation with contrastive learning as a core component, allowing the federated system to function without requiring clients to share any data features. Furthermore, its adaptable architecture can be used as an add-on module for existing unsupervised algorithms in federated settings. Experiments show that our model improves performance significantly (1.58--5.52pp) on five unsupervised algorithms.

\end{abstract}
\section{Introduction}
Most deep learning techniques assume unlimited access to data during training. However, this assumption does not hold in modern distributed systems, where data is stored at client nodes for privacy reasons~\cite{park2022knowledge,voigt2017eu}. For example, personal data stored on mobile devices cannot be shared with central servers, nor can patient records in hospital networks. \textit{Federated learning} is a new branch of collaborative technique to build a shared data model while securing data privacy; it is a method to run machine learning by involving multiple decentralized edge devices without exchanging locally bounded data~\cite{bonawitz2019towards,wang2021field}.

In federated systems, supervised methods have been used for a variety of downstream tasks such as object detection~\cite{liu2020fedvision}, image segmentation~\cite{sheller2018multi}, and person re-identification~\cite{zhuang2020performance}. The main challenge here is the data's decentralized and heterogeneous nature (i.e., non-IID setting), which obscures the global data distribution. To address this issue, several methods have been proposed, including knowledge distillation~\cite{zhuang2020performance}, control variates~\cite{karimireddy2020scaffold}, and contrastive learning~\cite{li2021model}. These methods necessitate that local clients have high-quality data labels.\looseness=-1

Nowadays, the need for \emph{unsupervised} federated learning is increasing to handle practical scenarios that lack data labels. This is the new frontier in federated learning. There have been a few new ideas; for instance, Zhang \textit{et al.} proposed FedCA, a model that uses local data features and external datasets to alleviate inconsistency in the representation space~\cite{zhang2020federated}. Wu \textit{et al.} proposed FCL, which exchanges encrypted local data features for privacy and introduces a neighborhood matching approach to cluster the decentralized data across clients~\cite{wu2021federated}. However, these approaches allow data sharing among local clients and raise privacy concerns.

We present \model{}, a new advancement in unsupervised learning on federated systems that learns semantic representation from local data and refines the central server's knowledge via \emph{knowledge distillation}. Unlike previous approaches, this model is privacy-preserving and does not rely on external datasets. The model introduces two novel considerations to the standard FedAvg~\cite{mcmahan2017communication} framework: \emph{local knowledge distillation} to train the network progressively based on local data and \emph{global knowledge distillation} to regularize data bias due to the non-IID setting. This two-sided knowledge flow distinguishes our model.

\begin{figure}[t!]
\centering
\begin{subfigure}[t]{0.47\textwidth}
      \centering\includegraphics[width=\textwidth]{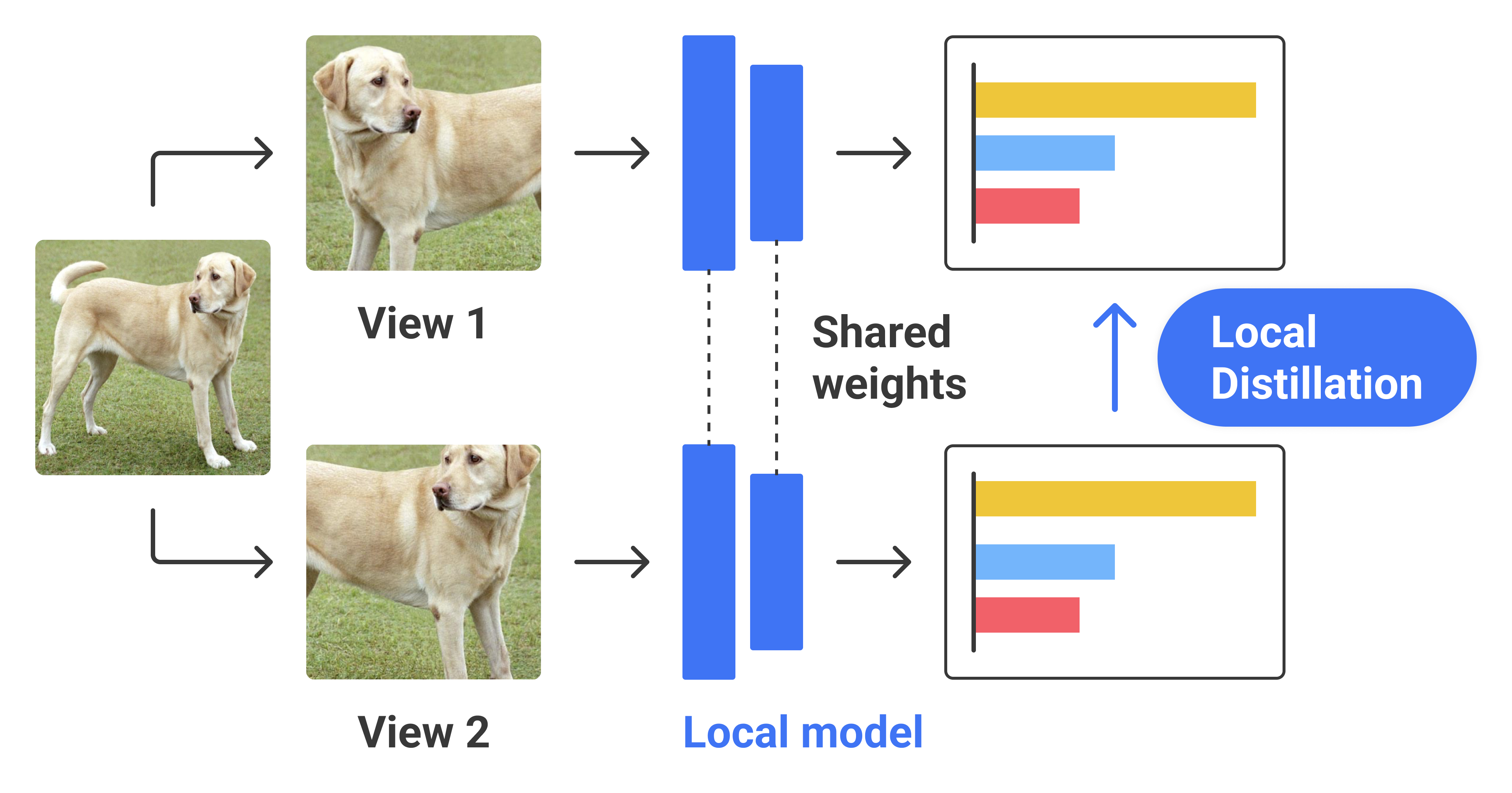}
      \subcaption{Local knowledge distillation}
      \label{fig:intro1}
\end{subfigure}
\begin{subfigure}[t]{0.47\textwidth}
      \centering\includegraphics[width=\textwidth]{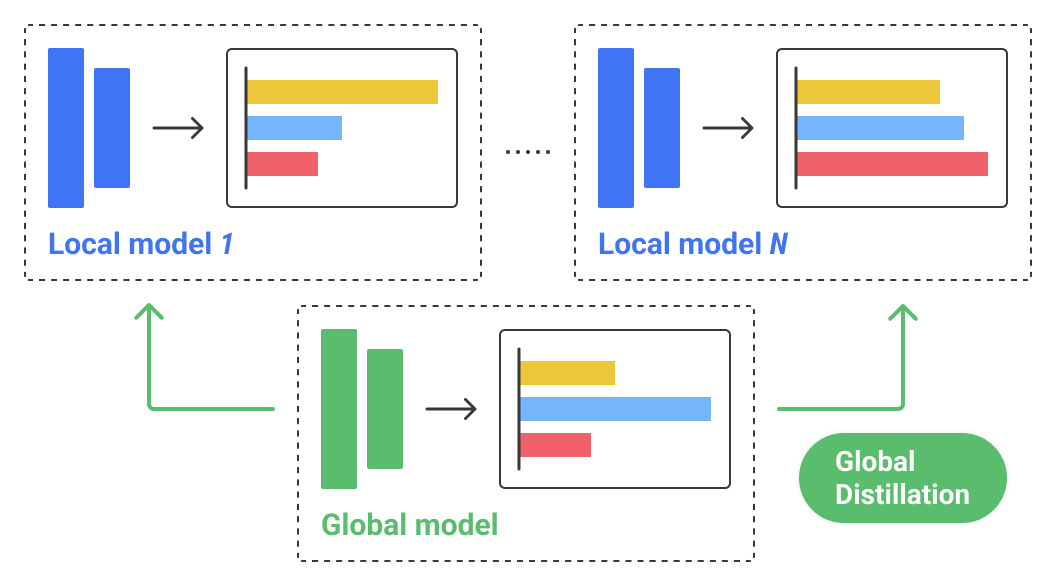} 
      \subcaption{Global knowledge distillation}
      \label{fig:intro2}
\end{subfigure}
\caption{
Illustration of two knowledge flows in \model{}: (a) local knowledge distillation progressively learns augmentation-invariant features, and (b) global knowledge distillation regularizes local models from bias.
} 
\label{fig:introduction}
\end{figure}

Local knowledge distillation (Fig.~\ref{fig:intro1}) maximizes the embedding similarity between two different views of the same data instance while minimizing that of other instances---this process is defined by the \textit{contrastive loss}. We designed an additional loss that relaxes the contrastive loss via soft labeling. Soft labels are computed as similarities between an anchor and randomly selected instances, called \textit{relationship vectors}. We minimize the distance between relationship vectors of two different views in order to transfer structural knowledge and achieve fast training speed---this process is modulated by the \textit{relational loss}.

Global knowledge distillation (Fig.~\ref{fig:intro2}) treats the sample representation passed by the global model as an alternative view that should be placed near the embedding of the local model. This process is also modulated by contrastive loss and relational loss. Concurrent optimization allows the model to learn semantic information while eliminating data bias through regularization. These objectives do not require additional communication rounds or costly computation. Moreover, they do not share sensitive local data or use external datasets.

\begin{enumerate}
    \item We propose an unsupervised federated learning algorithm, \model{}, that learns data representations via a unique two-sided knowledge distillation at local and global levels. 
    
    \item Two-sided knowledge distillation helps discover meaningful representation from local data while eliminating bias by using global knowledge.
    
    \item \model{} can be applied to extant algorithms to enhance performance by 1.58--5.52pp in top-1 accuracy and further enhance training speed.
    
    \item Unlike other unsupervised federated learning approaches, \model{} preserves privacy between clients and does not share data directly. It is also lightweight and does not require complex communication for sending data features.
    
    \item \model{} is open-sourced at \href{https://github.com/Sungwon-Han/FEDX}{https://github.com/Sungwon-Han/FEDX}.
\end{enumerate}

\section{Related Work}

\subsection{Unsupervised Representation Learning}

There are two common approaches to unsupervised representation learning. One approach is to use generative models like autoencoder~\cite{vincent2008extracting} and adversarial learning~\cite{radford2015unsupervised} that learn the latent representation by mimicking the actual data distribution. Another method is to use discriminative models with contrastive learning~\cite{gidaris2018unsupervised,park2021improving,xu2020comprehensive}. Contrastive learning approaches teach a model to pull the representations of the anchor and its positive samples (i.e., different views of the image) in embedding space, while pushing the anchor apart from negative samples (i.e., views from different images)~\cite{han2020mitigating,li2020prototypical}. 

In contrastive learning, SimCLR~\cite{chen2020simple} employs data augmentation to generate positive samples. MoCo~\cite{he2020momentum} introduces a momentum encoder and dynamic queue to handle negative samples efficiently. BYOL~\cite{grill2020bootstrap} reduces memory costs caused by a large number of negative samples. ProtoCL~\cite{li2020prototypical} uses prototypes to group semantically similar instances into local clusters via an expectation-maximization framework. However, under distributed and non-IID data settings, as in federated systems, these methods show a decrease in accuracy~\cite{zhang2020federated}.

\subsection{Federated Learning} 

Federated Averaging (FedAvg) by McMahan \textit{et al.} is a standard framework for supervised federated learning~\cite{mcmahan2017communication}. Several subsequent studies improved the local update or global aggregation processes of FedAvg. For instance, external dataset~\cite{zhao2018federated}, knowledge distillation~\cite{zhuang2020performance}, control variates~\cite{karimireddy2020scaffold,li2020federated}, and contrastive learning~\cite{li2021model} can be applied for better local update process. Similarly, global aggregation process can be improved via Bayesian non-parametric approaches~\cite{Wang2020Federated}, momentum updates~\cite{hsu2019measuring}, or normalization methods~\cite{wang2020tackling}. 

Unsupervised federated learning is more difficult to implement because no labels are provided and clients must rely on locally-defined pretext tasks that may be biased. This is a less explored field, with only a few methods proposed. FedCA~\cite{zhang2020federated} shares local data features and uses an external dataset to reduce the mismatch in representation space among clients. FCL~\cite{wu2021federated} encrypts the local data features before exchanging them. Because of the explicit data sharing, these methods raise new privacy concerns. We, on the other hand, consider a completely isolated condition that does not permit any local data sharing. FedU~\cite{zhuang2021collaborative} is another approach in the field that improves on the global aggregation method. It decides how to update predictors selectively based on the divergence of local and global models. Our model is orthogonal to FedU, and both concepts can be used in tandem.

\subsection{Knowledge Distillation}  

Knowledge distillation aims to effectively train a network (i.e., student) by distilling the knowledge of a pretrained network (i.e., teacher). Knowledge can be defined over the features at the intermediate hidden layers~\cite{komodakis2017paying,koratana2019lit}, logits at the final layer~\cite{hinton2015distilling}, or structural relations among training samples~\cite{park2019relational,tejankar2021isd,mitrovic2020representation}. Self-knowledge distillation uses the student network itself as a teacher network and progressively uses its knowledge to train the model~\cite{ji2021refine,kim2021self}. We leverage this concept to efficiently train the local model while preserving the knowledge of the global model. \model{} is the first-of-a-kind approach that uses the knowledge distillation concept for unsupervised federated learning.
\section{Model}

\subsection{Overview}
\textbf{Problem statement.~} Consider a federated system in which data can only be viewed locally at each client and cannot be shared outside. Our goal is to train a single unsupervised embedding model ${F}_\phi$ that maps data points from each client to the embedding space. Let us denote local data and model from client $m$ as $\mathcal{D}^m$ and $f_\theta^m$ respectively (i.e., $m \in \{1, ..., M\}$). The main objective for the global model ${F}_\phi$ is as follows:
\begin{align}
    \argmin_\phi \mathcal{L}(\phi) &= \sum_{m=1}^{M} {|\mathcal{D}^m| \over |\mathcal{D}|}\mathcal{L}_m(\phi), \nonumber \\
    \text{where } \mathcal{L}_m(\phi) &= \mathbb{E}_{\mathbf{x} \in \mathcal{D}^m}[l_m (\mathbf{x}; \phi)].
\end{align}
$\mathcal{L}_m$ represents the local objective in client $m$ and $l_m$ is the empirical loss objective of $\mathcal{L}_m$ over $\mathcal{D}^m$. For simplicity, we hereafter denote the local model $f_\theta^m$ at client $m$ and global model ${F}_\phi$ as $f^m$ and $F$. 

\begin{figure*}[t!]
\centerline{
      \includegraphics[width=\linewidth]{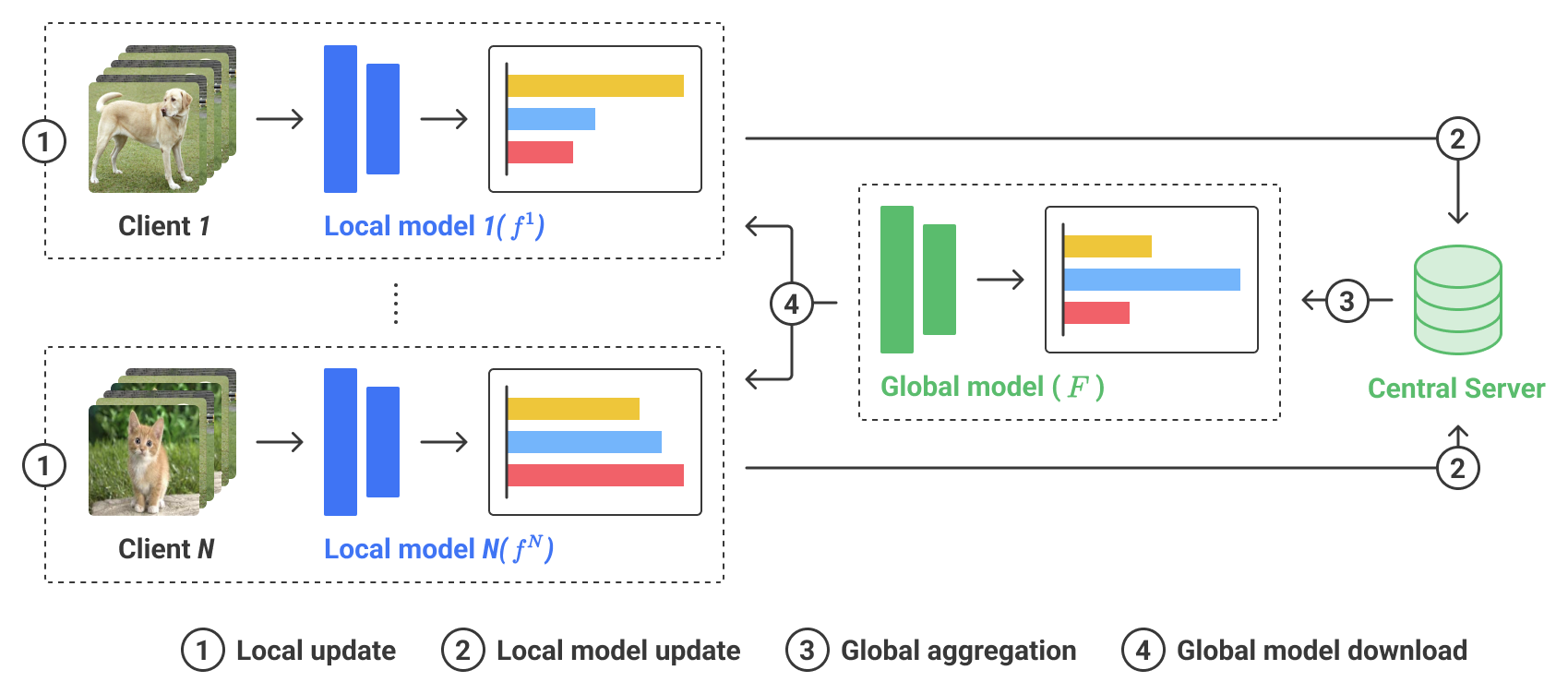}}
      \caption{Illustration of the FedAvg framework~\cite{mcmahan2017communication}, which is used as the base structure of many federated systems. \model{} modifies the local update process \textcircled{1}.
      } 
\label{fig:fedavg}
\end{figure*}

We use FedAvg~\cite{mcmahan2017communication} as the underlying structure, and the data flow is depicted in Fig.~\ref{fig:fedavg}. Four processes run in each communication round: \textsf{Process \textcircled{1} on local update} is when each local client trains a model $f^m$ with its data $\mathcal{D}^m$ for $E$ local epochs; \textsf{Process \textcircled{2} on local model upload} occurs when clients share the trained model weights with the server; \textsf{Process \textcircled{3} on global aggregation} occurs when the central server averages the received model weights and generates a shared global model $F$; \textsf{Process \textcircled{4} on global model download} is when clients replace their local models with the downloaded global model (i.e., averaged weights). These processes run for $R$ communication rounds.

\model{} modifies the \textsf{Process \textcircled{1}}  by redesigning loss objectives in order to distill knowledge at both the local and global scales. The following sections introduce the design components of our unsupervised federated learning model.

\subsection{Local Knowledge Distillation}
The first significant change takes place with local clients, whose goal is to learn meaningful representations from local data. Let us define a data pair; $\mathbf{x}_i$ and $\mathbf{\tilde{x}}_i$ be two augmented views of the same data instance. The \textit{local contrastive loss} $L_{\text{c}}^\text{local}$ learns semantic representation by maximizing the agreement between $\mathbf{x}_i$ and $\mathbf{\tilde{x}}_i$ while minimizing the agreement of views from different instances (i.e., negative samples). We showcase the proposed contrastive loss from two of the unsupervised representation learning methods as vanilla baselines.
\begin{figure}[t!]
\centering
\begin{subfigure}[t]{0.48\textwidth}
      \centering\includegraphics[width=\textwidth]{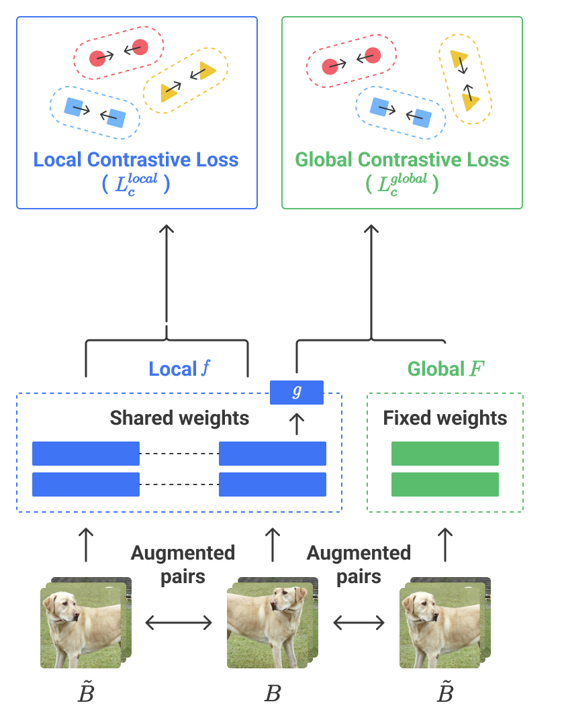}
      \subcaption{Two-sided contrastive loss}
      \label{fig:model_contrastive}
\end{subfigure}
\begin{subfigure}[t]{0.48\textwidth}
      \centering\includegraphics[width=\textwidth]{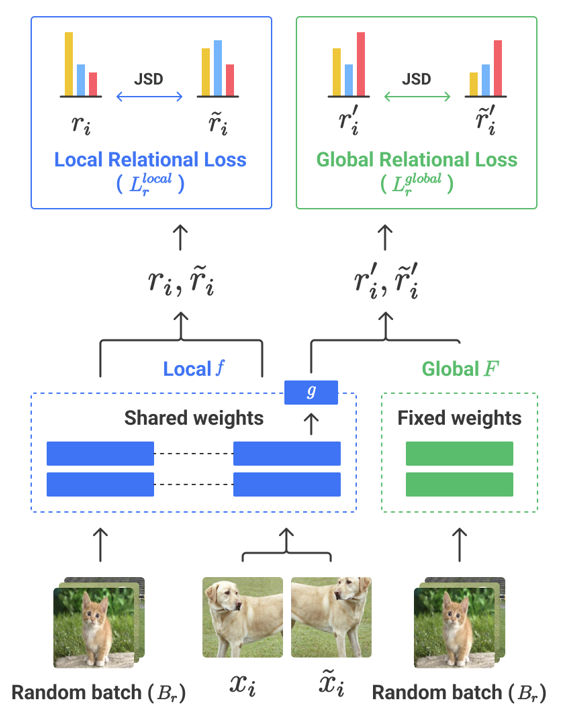} 
      \subcaption{Two-sided relational loss}
      \label{fig:model_relational}
\end{subfigure}
\caption{The overall architecture of \model{}, with the local model $f^{m}$, the projection head $h^{m}$, and the global model $F$ at local client $m$. Two-sided (a) contrastive loss and (b) relational loss enable the model to learn semantic information from local data while regularizing the bias by distilling knowledge from the global model. \model{} modifies the \textsf{process \textcircled{1} on local update} in Fig.~\ref{fig:fedavg}.} 
\label{fig:model}
\end{figure}

\begin{itemize}
\item[$\ast$] SimCLR~\cite{chen2020simple} utilizes a contrastive objective based on the InfoNCE loss~\cite{oord2018representation}. Given a batch $\mathcal{B}$ with size $n$ and its augmented version $\mathcal{\tilde{B}}$, each anchor has a single positive sample and considers all other ($2n-2$) data points to be negative samples. The following is the definition of this ($2n-1$)-way instance discrimination loss, where $\tau$ is the temperature used to control the entropy value and sim($\cdot$) is the cosine similarity function between two embeddings:
\begin{align}
    &L_{\text{c}}^\text{local} =  -\log {\text{exp}({\text{sim}(\mathbf{z}_i, \mathbf{\tilde{z}}_i) / \tau})  \over {\sum_{k \in (\mathcal{B} \cup \mathcal{\tilde{B}} - \{i\})} \text{exp}(\text{sim}(\mathbf{z}_i,  \mathbf{z}_k) / \tau)}}, \\
    &\text{where  }  \mathbf{z}_i = f^m (\mathbf{x}_i), \ \mathbf{\tilde{z}}_i = f^{m} (\mathbf{\tilde{x}}_i).\label{eq:contrastive_loss}  
\end{align}

\item[$\ast$] BYOL~\cite{grill2020bootstrap} does not train on negative samples. Instead, an asymmetric architecture is used to prevent the model from learning trivial solutions. The model $f^m$ with a prediction layer $g^m$ is trained to predict a view from the exponential moving average model $f^m_{\text{ema}}$. The loss is defined as follows:
\begin{align}
    &L_{\text{c}}^\text{local} = \norm[\Big]{ \mathbf{z}_{i} / \norm{ \mathbf{z}_{i}} -  \mathbf{\tilde{z}^{\text{ema}}}_i / \norm{\mathbf{\tilde{z}^{\text{ema}}}_i}}^2, \\
     &\text{where  }  \mathbf{z}_{i} = g^m \circ f^m (\mathbf{x}_i), \ \mathbf{\tilde{z}^{\text{ema}}}_i = f^m_{\text{ema}}(\mathbf{\tilde{x}}_i).\label{eq:byol_loss}
\end{align}
\end{itemize}

We consider another design aspect to help the model learn structural knowledge more effectively. Motivated by the concept of relational knowledge distillation~\cite{bhat2021distill,yang2022mutual}, structural knowledge represented as relations among samples is extracted from the local model and progressively transferred back to itself. This entails selecting a set of instances at random $\mathcal{B}_r$ and computing the cosine similarity between the embeddings of two different views $\mathbf{x}_i$, $\mathbf{\tilde{x}}_i$ and random instances $\mathcal{B}_r$. We then apply the softmax function to the similarity vector to compute relationship probability distributions $\mathbf{r}_i$ and $\mathbf{\tilde{r}}_i$ (Eq.~\ref{eq:relation}). In vector notation, the superscript $j$ represents the $j$-th component value of a given vector.
\begin{align}
    \mathbf{r}_i^j  = \frac{\exp(\text{sim}(\mathbf{z}_i, \mathbf{z}_j)/\tau)}{\sum_{k \in \mathcal{B}_r} \exp(\text{sim}(\mathbf{z}_i, \mathbf{z}_k)/\tau)},\ \ \mathbf{\tilde{r}}_i^j  = \frac{\exp(\text{sim}(\mathbf{\tilde{z}}_i, \mathbf{z}_j)/\tau)}{\sum_{k \in \mathcal{B}_r} \exp(\text{sim}(\mathbf{\tilde{z}}_i, \mathbf{z}_k)/\tau)} \label{eq:relation}
\end{align}

The above concept, \emph{local relational loss}, is defined as the Jensen-Shannon divergence (JSD) between two relationship probability distributions $\mathbf{r}_i$ and $\mathbf{\tilde{r}}_i$ (Eq.~\ref{eq:relational_loss}). Minimizing the discrepancy between two distributions make the model to learn structural knowledge invariant to data augmentation. In contrastive learning with soft targets, this divergence loss can also be interpreted as relaxing the InfoNCE objective.
\begin{align}
    L_{r}^\text{local} = \frac{1}{2} \text{KL}(\mathbf{r}_i \Vert \mathbf{r}_i^{\text{target}}) + \frac{1}{2} \text{KL}(\mathbf{\tilde{r}}_i \Vert \mathbf{r}_i^{\text{target}}),\text{ where }     \mathbf{r}_i^{\text{target}} = \frac{1}{2}(\mathbf{r}_i + \mathbf{\tilde{r}}_i) \label{eq:relational_loss}
\end{align}
The total loss term for local knowledge distillation is given in Eq.~\ref{eq:total_loss}: 
\begin{align}
    L_{\text{local-KD}} = L_{c}^\text{local} + L_{r}^\text{local}. \label{eq:total_loss}
\end{align}

\subsection{Global Knowledge Distillation}
The second major change is to regularize the bias contributed by the inconsistency between local and overall data distribution. The inconsistency addresses the issue of decentralized non-IID settings, where local clients are unaware of global data distribution. Training the local model $f^{m}$ will be suboptimal in this case because the local update process becomes biased towards local minimizers~\cite{zhang2020federated}. Such data inconsistency among local clients can be resolved by distilling knowledge on a global scale. \looseness=-1

We consider two kinds of losses: \textit{global contrastive loss} and \textit{global relational loss}. Because the global model simply aggregates model weights at the local clients in FedAvg, we can think of the sample's embedding from the global model as an alternate view of the same data instance. The global contrastive loss maximizes the agreement between the views of the local and global models from the same instance while minimizing that of all other views from different instances.\looseness=-1

Each communication round assumes that the central server sends a fixed set of averaged model weights (i.e., global model $F$) to the client. The batch $\mathcal{B}$ and its augmented version $\mathcal{\tilde{B}}$ are then used to train the local model $f^m$ as in Eq.~\ref{eq:contrastive_loss_global} with the InfoNCE loss. To match the embedding space between the local and global models, we consider an additional prediction layer $h^{m}$ on top of local models. Similar method has been used in~\cite{chen2021exploring,grill2020bootstrap}.  \looseness=-1
\begin{align}
    L_{c}^\text{global} = &-\log {\text{exp}({\text{sim}(\mathbf{z}^l_i, \mathbf{\tilde{z}}^g_i) / \tau})  \over {\sum\limits_{k \in (\mathcal{B} - \{i\})} \text{exp}(\text{sim}(\mathbf{z}^l_i, \mathbf{z}^l_k) / \tau) + \sum\limits_{k \in (\mathcal{\tilde{B}} - \{i\})}} \text{exp}(\text{sim}(\mathbf{z}^l_i, \mathbf{z}^g_k) / \tau)}, \nonumber \\
    \text{where  }  \mathbf{z}^l_i &= h^{m} \circ f^{m} (\mathbf{x}_i), \ \mathbf{\tilde{z}}^l_i = h^{m} \circ f^{m} (\mathbf{\tilde{x}}_i), \  \mathbf{z}^g_i = F (\mathbf{x}_i), \  \mathbf{\tilde{z}}^g_i = F (\mathbf{\tilde{x}}_i). \label{eq:contrastive_loss_global}
\end{align}

We introduce the \textit{global relational loss} on top of the global contrastive loss. This loss is defined in the same way as the local relational loss (Eq.~\ref{eq:relational_loss}), but it includes global model embeddings. It regularizes the model by penalizing any mismatch between two augmented views over the global embedding space after the prediction layer $h^{m}$. As a result, the model maintains its local knowledge based on local data while learning augmentation-invariant knowledge using the global contrastive loss. \looseness=-1

Given two different views $\mathbf{x}_i$, $\mathbf{\tilde{x}}_i$ and random instances $\mathcal{B}_r$, the relationship probability distributions for global relational loss, $\mathbf{r}'_i$ and $\mathbf{\tilde{r}}'_i$, are defined (Eq.~\ref{eq:global_relation}). We again adopt the JS divergence between two relationship probability vectors $\mathbf{r}'_i$ and $\mathbf{\tilde{r}}'_i$ as the global relational loss (Eq.~\ref{eq:global_relational_loss}).
\begin{align}
    \mathbf{r}_{i}'^j &= \frac{\exp(\text{sim}(\mathbf{z}_i^l, \mathbf{z}_j^g)/\tau)}{\sum_{k \in \mathcal{B}_r} \exp(\text{sim}(\mathbf{z}_i^l, \mathbf{z}_k^g)/\tau)},\ \ \mathbf{\tilde{r}}_{i}'^{j}  = \frac{\exp(\text{sim}(\mathbf{\tilde{z}}_i^l, \mathbf{z}_j^g)/\tau)}{\sum_{k \in \mathcal{B}_r} \exp(\text{sim}(\mathbf{\tilde{z}}_i^l, \mathbf{z}_k^g)/\tau)} \label{eq:global_relation} \\
    L_{r}^\text{global} &= \frac{1}{2} \text{KL}(\mathbf{r}'_i \Vert \mathbf{r}_{i}'^{\text{target}}) + \frac{1}{2} \text{KL}(\mathbf{\tilde{r}}'_i \Vert \mathbf{r}_{i}'^{\text{target}}),\text{ where }     \mathbf{r}_{i}'^{\text{target}} = \frac{1}{2}(\mathbf{r}'_i + \mathbf{\tilde{r}}_{i}') \label{eq:global_relational_loss}
\end{align}

The total loss for global knowledge distillation is given in Eq.~\ref{eq:total_loss_global}. The overall model then combines losses from knowledge distillation at the local and global levels, as shown in Eq.~\ref{eq:overall_loss}. The detailed algorithm is described in the appendix.
\begin{align}
    &L_{\text{global-KD}} = L_{c}^\text{global} + L_{r}^\text{global} \label{eq:total_loss_global} \\
    &L_{\text{total-KD}} = L_{\text{local-KD}} + L_{\text{global-KD}} \label{eq:overall_loss}
\end{align}

\section{Experiment}
Using multiple datasets, we compared the performance of our model to other baselines and investigated the role of model components and hyperparameters. We also used embedding analysis to examine how the proposed model achieves the performance gain. Finally, we applied the model in a semi-supervised setting.

\subsection{Performance Evaluation}
\noindent
\textbf{Data settings.~} Three benchmark datasets are used. CIFAR-10~\cite{krizhevsky2009learning} contains 60,000 images of 32$\times$32 pixels from ten classes that include airplanes, cats, and dogs. SVHN~\cite{netzer2011reading} contains 73,257 training images and 26,032 test images with small cropped digits of of 32$\times$32 pixels. F-MNIST~\cite{xiao2017fashion} contains 70,000 images of 28$\times$28 pixels from ten classes, including dresses, shirts, and sneakers. \looseness=-1

We used the Dirichlet distribution to enforce the non-IID property of local clients. Let $Dir_{N}(\beta)$ denote the Dirichlet distribution with $N$ clients and $\beta$ as the concentration parameter. We take a sample $p_{k,j}$ from $Dir_{N} (\beta)$ and assign class $k$ to client $j$ based on the sampled proportion $p_{k,j}$. With this data allocation strategy, each client will be assigned a few data samples for each class (or even none) to ensure bias. By default, $N$ and $\beta$ are to 10 and 0.5, respectively, similar to other research~\cite{li2021model}.\bigskip

\noindent
\textbf{Implementation details.~} 
The model was trained for 100 communication rounds, with 10 local epochs in each round. The ResNet18 backbone~\cite{he2016deep} and the SGD optimizer with a learning rate of 0.01 were used. SGD weight decay was set to 1e-5, SGD momentum was set to 0.9, and batch size was set to 128. For all objectives, the temperature $\tau$ was set as 0.1. Augmentations included random crop, random horizontal flip, and color jitter. We used four A100 GPUs. \looseness=-1 \bigskip

\noindent
\textbf{Baselines.~} We implemented five baselines: (1) FedSimCLR based on SimCLR~\cite{chen2020simple}, (2) FedMoCo based on MoCo~\cite{he2020momentum}, (3) FedBYOL based on BYOL~\cite{grill2020bootstrap}, and (4) FedProtoCL based on ProtoCL~\cite{li2020prototypical}. These are unsupervised models that are built on top of FedAvg~\cite{mcmahan2017communication}. The final baseline (5) FedU~\cite{zhuang2021collaborative} is built over FedBYOL and downloads a global model by divergence-aware module (see \textsf{process \textcircled{4}} in Figure~\ref{fig:fedavg}). For a fair comparison, we applied the same experimental settings on these baselines, including the backbone network, optimizer, augmentation strategy, number of local epochs, and communication rounds. We used the original implementations and hyper-parameter settings for FedU. Unless otherwise specified, we refer to FedSimCLR as the representative baseline in the remainder of this section. \bigskip

\begin{table}[t!]
\caption{Performance improvement with \model{} on classification accuracy over three datasets. Both the final round accuracy and the best accuracy show that our model brings substantial improvement for all baseline algorithms.}
\centering
{

\begin{tabular}{l|cc|cc|cc}
\toprule
\multirow{2}{*}{Method} & \multicolumn{2}{c|}{CIFAR-10} &  \multicolumn{2}{c|}{SVHN} & \multicolumn{2}{c}{F-MNIST} \\ 
                      & ~~ Last ~~  & ~~ Best ~~  & 
                      ~~ Last ~~ & ~~ Best ~~  & ~~ Last ~~ & ~~ Best ~~ \\ \midrule
FedSimCLR             &   51.31     &  52.88      &   75.19    &  76.50       &  77.66      &    79.44   \\
+ \model{}           & \textbf{56.88} & \textbf{57.95} &   \textbf{77.19} & \textbf{77.70} &\textbf{81.98} & \textbf{82.47}   \\ \hline
FedMoCo               &	56.74 & 57.82 &	70.69 &  70.99& 82.31 & 83.58 \\
+ \model{}            	&  \textbf{58.23} & \textbf{59.43}  & \textbf{73.57} & \textbf{73.92}  & \textbf{83.62}  & \textbf{84.65}  \\ \hline
FedBYOL              &	52.24  & 53.14	 	&65.95 & 67.32		&81.45 & 82.37 \\
+ \model{}           	&\textbf{56.49} & \textbf{57.79}& 	\textbf{68.94} & \textbf{69.05} &  \textbf{83.18} 	 & \textbf{84.30}   \\ \hline
FedProtoCL          	& 51.33 & 52.12 & 49.85 & 50.19	&  81.76 & \textbf{83.57}   \\
+ \model{}             &  \textbf{55.36}   & \textbf{56.76}   &   \textbf{69.31}    &   \textbf{69.75}   &    \textbf{82.74}    &   83.34    \\ \hline
FedU                   &  50.79   & 50.79         &   66.02  &  66.22     &  80.59     &   82.03         \\
+ \model{}                &  \textbf{56.15}   &     \textbf{57.26}  &     \textbf{68.13}    &  \textbf{68.39}   &   \textbf{83.73}    &   \textbf{84.12}    \\ \bottomrule
\end{tabular}
}
\label{tab:main_result}
\end{table}

\begin{figure}[t!]
\centering
\begin{subfigure}[t]{0.96\textwidth}
      \centering\includegraphics[width=0.99\textwidth]{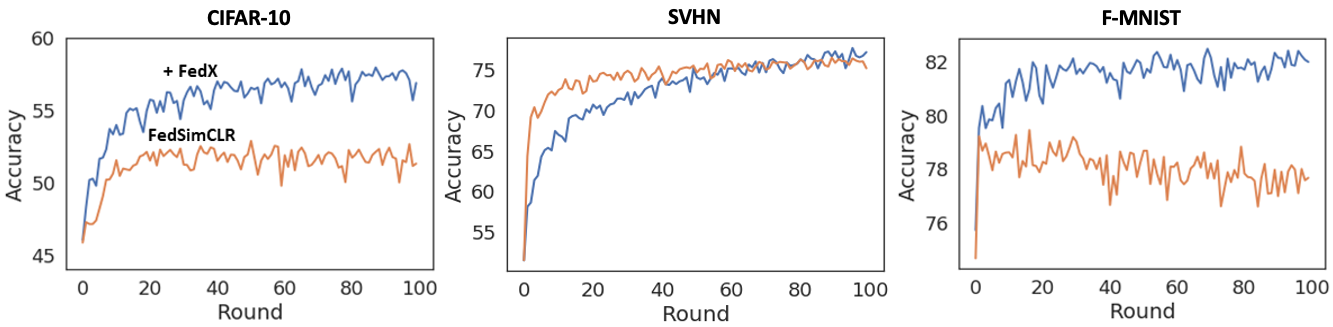}
      \subcaption{Performance gain on FedSimCLR}
      \label{fig:fedsimclr_round}
\end{subfigure}
\begin{subfigure}[t]{0.96\textwidth}
     \centering\includegraphics[width=0.99\textwidth]{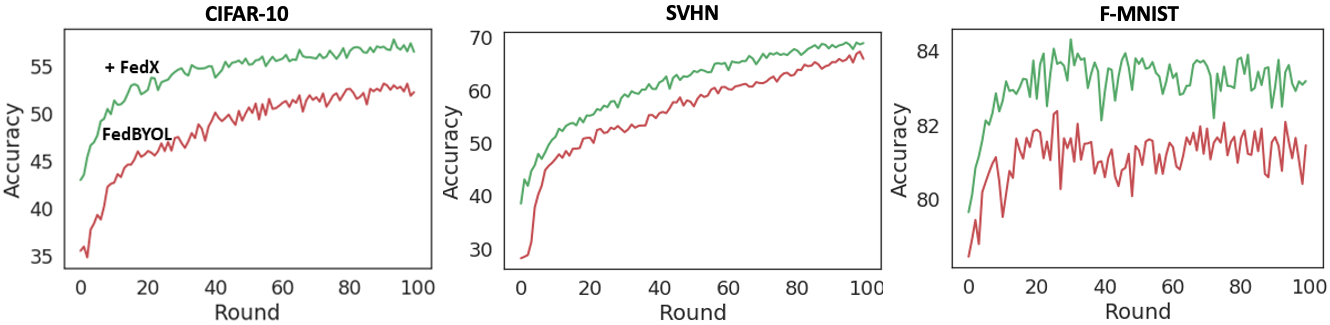}
      \subcaption{Performance gain on FedBYOL}
      \label{fig:fedbyol_round}
\end{subfigure}
\caption{Performance comparison between two vanilla baselines (i.e., FedSimCLR and FedBYOL) and \model{}-enhanced versions over communication rounds. \model{} helps models outperform in all three benchmark datasets and continues to bring advantage with increasing communication rounds.} 
\label{fig:round}
\end{figure}

\noindent
\textbf{Evaluation.~}
All models were compared using the linear evaluation protocol, which is a method for training a linear classifier on top of representations~\cite{zhang2020federated,zhuang2021collaborative}. We freeze the backbone network of each trained model after training. Then, for the next 100 epochs, a new classifier is appended and trained with ground-truth labels. The top-1 classification accuracy over the test set is reported as an evaluation metric. \bigskip

\noindent
\textbf{Results.~}
Table~\ref{tab:main_result} summarizes the performance comparison, where \model{} brings meaningful performance improvements over the baseline algorithms. On average, our model improves CIFAR-10 by 4.29 percent points (pp), SVHN by 5.52pp, and F-MNIST by 1.58pp across all baselines. One exception is F-MNIST, where FedProtoCL by itself has a slightly higher best accuracy. However, adding \model{} still contributes to improved final round accuracy, implying that the model has good training stability. \looseness=-1

We then examine how quickly the model improves baselines across the various communication rounds. Figure~\ref{fig:round} shows the trajectory for two example baselines on FedSimCLR and FedBYOL.\footnote {Results for other baselines are presented in the Appendix.} These plots confirm that model-enhanced models outperform vanilla baselines; most plots show this benefit early in the communication rounds. We see that local bias can degrade the performance of a baseline model during the early training phase in some cases (see the F-MNIST case in Figure~\ref{fig:fedsimclr_round}). This is most likely due to the biased contrastive objective caused by locally sampled negatives. In contrast, adding \model{} prevents such deterioration and even continues to improve accuracy as communication rounds increase. \looseness=-1

\subsection{Component Analyses}

\begin{figure}[t!]
 \centering\includegraphics[width=1\textwidth]{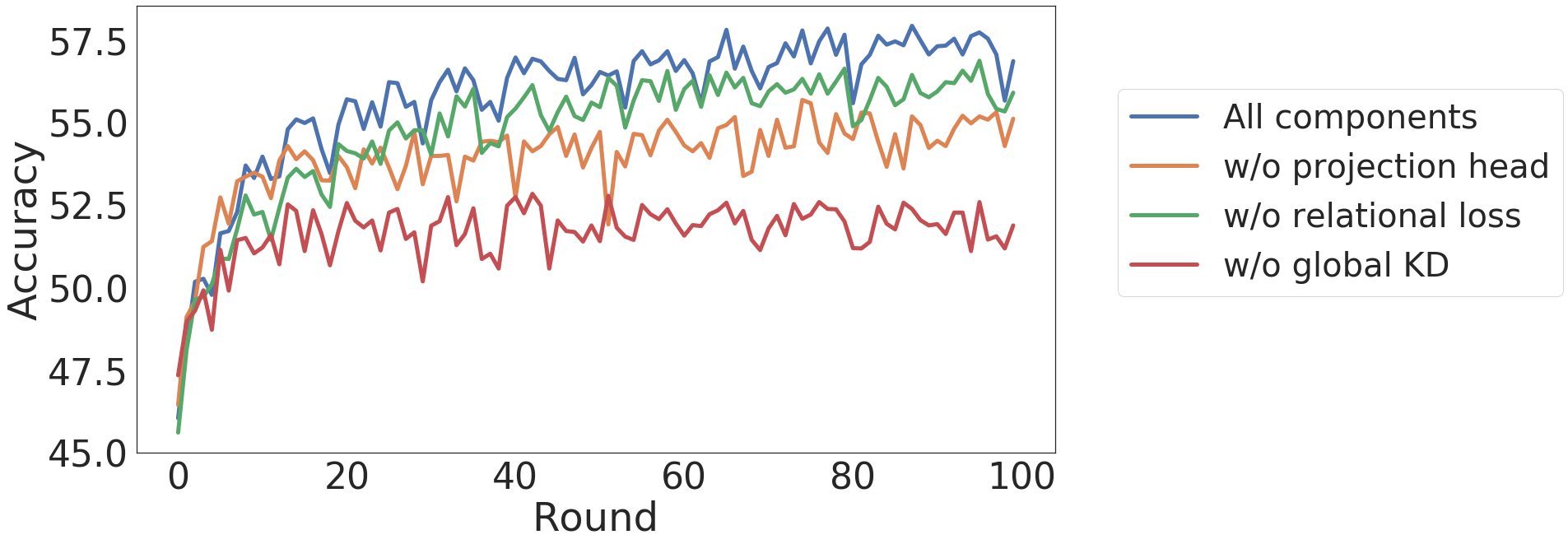}
\caption{Performance comparison of ablations over communication rounds for CIFAR-10. Removing any module leads to performance degradation. Ablation on contrastive loss $L_c$ showed the best accuracy of 35.13\% and hence excluded. \looseness=-1} 
\label{fig:ablation_study}
\end{figure}

\noindent
\textbf{Ablation study.~}\model{} used learning objectives at the local and global levels separately, with two types of losses: contrastive loss and relational loss. In this section, we look at ablations by removing each learning objective or loss component and testing the added value of each design choice to overall performance. Figure~\ref{fig:ablation_study} plots the performance comparison of different ablations across the communication round. The complete model has the highest accuracy, implying that removing any component reduces performance. It also confirms the importance of a global knowledge distillation objective. 

\model{} used global knowledge distillation to convey global model knowledge and regularize the local bias caused by the inconsistency between local and overall data distribution. Several studies in supervised settings have addressed a similar challenge using extra regularization or gradient update processes. We replaced the global knowledge distillation loss ($L_{\text{global-KD}}$ -- Eq.~\ref{eq:total_loss_global}) with extant strategies, such as FedProx~\cite{li2020fedprox} or SCAFFOLD~\cite{karimireddy2020scaffold} and verified its efficacy. The performance comparison of different ablations across three benchmark datasets is summarized in Table~\ref{tab:further_ablation}. The findings imply that our global knowledge distillation technique is more effective than alternative designs. \bigskip

\begin{table}[t!]
\caption{Ablation results with different global-scale regularization methods. The proposed global knowledge distillation performs the best among them.}
\centering
\begin{tabular}{l|cccccc}
\toprule
\multirow{2}{*}{Method} & \multicolumn{2}{c}{CIFAR-10} & \multicolumn{2}{c}{SVHN} & \multicolumn{2}{c}{F-MNIST} \\
                        &  ~~Last~~         &  ~~Best~~          &  ~~Last~~       & ~~Best~~        &  ~~Last~~         &  ~~Best~~       \\ \midrule
$L_{\text{local-KD}}$ only  &   51.89     &  52.85        &     76.64        &          77.20  &      79.79        &  80.42             \\
$L_{\text{local-KD}}$ + SCAFFOLD     &   52.73     &    53.20     &     75.18        &     75.52       & 79.45             &         80.36       \\
$L_{\text{local-KD}}$ + FedProx                 &  52.48             &     53.34         &    \textbf{77.43}          &   \textbf{77.79}  &       79.83               &           80.24     \\
$L_{\text{local-KD}}$ + $L_{\text{global-KD}}$  &  \textbf{56.88} &  \textbf{57.95}    &    77.19   & 77.70           &        \textbf{81.98}       &    \textbf{82.47}          \\ \bottomrule
\end{tabular}
\label{tab:further_ablation}
\end{table}

\begin{table}[t!]
\caption{Analysis of accuracy on CIFAR-10 over varying hyper-parameters indicates \model{} consistently enhances the baseline performance.}
\centering
\begin{subtable}[t]{0.48\textwidth}
\centering
\caption{Effect of the data size $|\mathcal{D}|$}
\begin{tabular}{c|cc|cc}
\toprule
\multirow{2}{*}{Data size} & \multicolumn{2}{c|}{Baseline} & \multicolumn{2}{c}{Baseline+\model{}} \\
 &  ~Last~   &   ~Best~  & ~~Last~   &   ~Best~   \\ \midrule
10\%                       & 46.80         & 47.37         & 51.03         & 53.96         \\
25\%                       & 48.42         & 49.79         & 52.84         & 54.45         \\
50\%                       & 51.17         & 52.04         & 54.62         & 55.85         \\
100\%                      & 51.31         & 52.88         & 56.88         & 57.95         \\ \bottomrule
\end{tabular}
\label{tab:table1_a}
\end{subtable}
\begin{subtable}[t]{0.48\textwidth}
\centering
\caption{Effect of the client count $N$}
\begin{tabular}{c|cc|cc}
\toprule
\multirow{2}{*}{Clients \#} & \multicolumn{2}{c|}{Baseline} & \multicolumn{2}{c}{Baseline+\model{}} \\
 & ~Last~   &   ~Best~  &  ~~Last~   &   ~Best~    \\  \midrule
 5 & 52.87 & 53.87 & 58.55 & 58.55 \\
 10 & 51.31 & 52.88 & 56.88 & 57.95 \\
 15 & 52.31 & 53.06 & 55.12 & 56.82 \\
 20 & 50.70 & 52.89 & 56.56 & 56.56 \\ \bottomrule
\end{tabular}
\label{tab:table1_b}
\end{subtable}
\begin{subtable}[t]{0.95\textwidth}
\centering
\caption{Effect of the communication round count $R$}
\begin{tabular}{c|cc|cc}
\toprule
\multirow{2}{*}{Communication round} & \multicolumn{2}{c|}{Baseline} & \multicolumn{2}{c}{Baseline+\model{}} \\
 &  ~~Last~~   &   ~~Best~~  &  ~~Last~~   &   ~~Best~~ \\  \midrule
20 & 52.01 & 52.80 & 56.97 & 56.97 \\ 
50 & 51.95 & 53.53 & 57.29 & 57.29 \\
100 & 51.31 & 52.88 & 56.88 & 57.95 \\
200 & 52.79 & 53.23 & 57.35 & 57.58 \\\bottomrule
\end{tabular}
\label{tab:table1_c}
\end{subtable}
\label{tab:robustness}
\end{table}

\noindent
\textbf{Robustness test.~} The model's robustness is then tested by varying key hyperparameters in different simulation settings. This allows us to test the system in difficult scenarios, such as (a) when each client is only allowed to hold a small amount of data (i.e., data size $|\mathcal{D}|$), (b) when more clients participate in the federated system (i.e., client count $N$), and (c) when communication with the central server becomes limited and costly (i.e., the number of communication rounds $R$)~\cite{wang2021field}. We test how our model performs under these scenarios in Table~\ref{tab:robustness}. We note that when varying the communication rounds $R$, we also changed the number of local epochs $E$ accordingly such that $R\times E = 1000$.

The table summarizes the effect of each hyperparameter for the baseline model (FedSimCLR) and the \model{}-enhanced model. We make several observations. First, reducing the data size $|\mathcal{D}|$ degrades performance. The drop, however, is not severe and remains nearly 5pp drop even when clients only hold 10\% of the data. Second, increasing the number of clients $N$ will add complexity and degrade performance. However, when $N$ increases from 10 to 20, the drop is only marginal near 1pp. Third, while increasing communication rounds generally provides additional benefits, the gain appears to be marginal after some rounds, as shown in the example. Regardless of these changes, \model{} consistently leads to nontrivial improvements over baseline.

\subsection{Analysis of the Embedding Space} 
We next quantitatively examine the embedding space characteristics to see how well \model{} distills global knowledge into the local model and encodes data semantic structure. We calculated the angle difference between the normalized embeddings passed by local model $f$ and global model $F$ as a quality metric: \looseness=-1
\begin{align}
    \text{Angle}(\mathbf{x}) = \arccos{(\text{sim}(f(\mathbf{x}), F(\mathbf{x})))}, \label{eq:angle}
\end{align}
where $\mathbf{x}$ is an instance from the test data $\mathcal{D}_{\text{test}}$ and sim($\cdot$) is the cosine similarity function. It should be noted that a larger angle represents more significant deviance in the embedding distributions of the two models.

\begin{figure}[t!]
\centering
\begin{subfigure}[t]{0.45\textwidth}
      \centering\includegraphics[width=\textwidth]{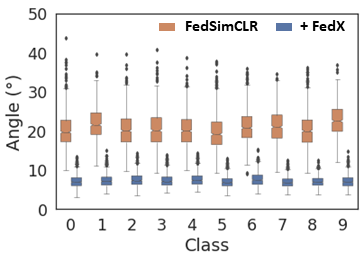}
      \subcaption{Box plot on local vs. global models}
      \label{fig:inner_class}
\end{subfigure}
\begin{subfigure}[t]{0.48\textwidth}
      \centering\includegraphics[width=\textwidth]{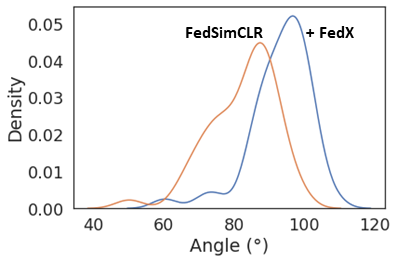} 
      \subcaption{Histogram on inter-class difference}
      \label{fig:inter_class}
\end{subfigure}
\caption{Embedding analysis of baseline and \model{}-enhanced models on CIFAR-10 comparing the angle difference of the embedded features.} 
\label{fig:qualitative_analysis}
\end{figure}

Figure~\ref{fig:inner_class} visualizes, for each of the ten classes in CIFAR-10, the angle difference between the embedding of each item between the local model and the global model computed by Eq.~\ref{eq:angle}. Compared to the baseline (FedSimCLR), \model{}-enhanced model reports a remarkably lower angle difference between the local and global models. This indicates that the local model can learn the refined knowledge of the global model through knowledge distillation.

When it comes to the embedding space of different class items, it is best to have a large gap. Given $\mathcal{D}_{\text{test}}^c$ as a set of instances from class $c$, we can compute a representative class prototype by averaging embeddings from $\mathcal{D}_{\text{test}}^c$ (Eq.~\ref{eq:prototype}). Then, the inter-class angle difference can be defined between any pair of class prototypes (Eq.~\ref{eq:interclassangle}). Figure~\ref{fig:inter_class} plots the histogram of the inter-class angle difference of every class pair, showing that \model{}-enhanced models have larger angles of 93.15$^{\circ}$ on average than the baseline model of 82.36$^{\circ}$. This demonstrates that our model can better discriminate between different class items. 
\begin{align}
    \mathbf{z}_c &= {1 \over |\mathcal{D}^c_{\text{test}}|} \sum_{\mathbf{x} \in \mathcal{D}^c_{\text{test}}}{f(\mathbf{x})} \label{eq:prototype} \\
    \text{Angle}(c_i, c_j) &= \arccos (\text{sim}(\mathbf{z}_{c_i}, \mathbf{z}_{c_j})) \label{eq:interclassangle}
\end{align}

\subsection{Extension to Semi-Supervised Settings}  
Finally, as a practical extension, consider a scenario in which each client has a small set of partially labeled data. This may be a more natural setting in many real-world federated systems~\cite{zhuang2021collaborative}. To convert our model to a semi-supervised setting, we first trained it without supervision and then fine-tuned it with an additional classifier on labeled data for an additional 100 epochs. For fine-tuning, an SGD optimizer with a learning rate of 1e-3 was used.  \looseness=-1

Table~\ref{tab:semi} shows the performance results on CIFAR-10 in the semi-supervised setting with varying label ratios of 1\%, 5\%, and 10\%. As expected, increasing the labeling ratio from 1\% to 5\% brings an immediate performance gain. \model{}-enhanced models outperform most cases in the semi-supervised setting for multiple baselines. Only minor exceptions can be seen with a 1\% labeling rate, where our model performs similarly to the baseline. Our model, on the other hand, benefits more quickly from increasing the label ratio and can learn the data representation from distributed local clients.

\begin{table}[t!]
\caption{Classification accuracy in a semi-supervised setting on CIFAR-10. \model{} enhances the baseline performance even with a small set of labels.}
\centering
\begin{tabular}{c|cc|cc|cc|cc|cc}
\toprule
\multirow{2}{*}{Label Ratio} & \multicolumn{2}{c|}{FedSimCLR}       & \multicolumn{2}{c|}{FedMoCo}         & \multicolumn{2}{c|}{FedBYOL}         & \multicolumn{2}{c|}{FedProtoCL}      & \multicolumn{2}{c}{FedU}             \\ \cmidrule{2-11} 
                             & \multicolumn{1}{c}{Vanilla} & \model{}  & \multicolumn{1}{c}{Vanilla} & \model{}  & \multicolumn{1}{c}{Vanilla} & \model{}  & \multicolumn{1}{c}{Vanilla} & \model{}  & \multicolumn{1}{c}{Vanilla} & \model{}  \\ \midrule
1\%                          & \multicolumn{1}{c}{21.37}   & \textbf{23.33} & \multicolumn{1}{c}{23.02}   & \textbf{25.18} & \multicolumn{1}{c}{18.10}        & \textbf{21.86} & \multicolumn{1}{c}{\textbf{18.44}}        & 18.17 & \multicolumn{1}{c}{\textbf{21.41}}   & 21.23 \\
5\%                          & \multicolumn{1}{c}{30.68}   & \textbf{35.86} & \multicolumn{1}{c}{34.24}   & \textbf{37.63} & \multicolumn{1}{c}{29.77}        & \textbf{34.48} & \multicolumn{1}{c}{19.64}        & \textbf{26.66} & \multicolumn{1}{c}{32.19}   & \textbf{35.41} \\
10\%                         & \multicolumn{1}{c}{31.14}   & \textbf{39.40}  & \multicolumn{1}{c}{38.15}   & \textbf{39.32} & \multicolumn{1}{c}{32.23}        & \textbf{37.89} & \multicolumn{1}{c}{22.90}        & \textbf{27.54} & \multicolumn{1}{c}{34.51}   & \textbf{37.51} \\ \bottomrule
\end{tabular}
\label{tab:semi}
\end{table}

\section{Conclusion}
This work presented the first-of-its-kind unsupervised federated learning approach called \model{}. We elaborate the local update process of the common federated learning framework and the model does not share any data directly across local clients. Its unique two-sided knowledge distillation can efficiently handle data bias in a non-IID setting while maintaining privacy. It is straightforward and does not require any complex communication strategy.

The substantial performance gain of \model{} shows great potential for many future applications. For example, distributed systems with strict data privacy and security requirements, such as learning patterns of new diseases across hospital data or learning tending content in a distributed IoT network, can benefit from our model. Unsupervised learning is facilitated even when local clients lack data labels and contain heterogeneous data. This versatile and robust trait makes unsupervised learning the new frontier in federated systems. We hope that our technique and implementation details will be useful in tackling difficult problems with decentralized data.

\section*{Acknowledgements}
We thank Seungeon Lee and Xiting Wang for their insights and discussions on our work. This research was supported by the Institute for Basic Science (IBS-R029-C2, IBS-R029-Y4), Microsoft Research Asia, and Potential Individuals Global Training Program (2021-0-01696) by the Ministry of Science and ICT in Korea. \looseness=-1

%
%
\bibliographystyle{splncs04}
\bibliography{eccv2022submission}

\newpage
\section{Appendix}
\subsection{Code Release \& Implementation details}
\model{} is open-sourced at \href{https://github.com/Sungwon-Han/FEDX}{https://github.com/Sungwon-Han/FEDX}. Algorithm 1 shows the overall training process.

\begin{algorithm}[h!]
\caption{Local update process of \model{}}
\begin{flushleft}
\vspace{-2mm}
\textbf{Input:} the number of local epochs $E$, global model $F$, local model $f^{m}$, projection head $h^{m}$, local dataset $\mathcal{D}^{m}$, and temperature $\tau$  \\
\textbf{Output:} a trained local model $f^{m}$ and a projection head $h^{m}$ \\ \vspace{3mm}
In each client $m$, \\
$f^{m} \leftarrow F$ \hfill \tcp{Replace the local model with the global model}
$F \leftarrow F.detach()$ \hfill \tcp{Fix the global model for a given communication round} \vspace{-3mm}
\end{flushleft}
    \For{epoch $e \in \{1,2,..., E\}$}
    {
        \For{batch $\mathcal{B}, \mathcal{B}_r \in \mathcal{D}^i$}
        {
            $\mathcal{\tilde{B}} \leftarrow$ Augment($\mathcal{B}$) \\
             \For{$\mathbf{x}_i \in \mathcal{B}$, $\mathbf{\tilde{x}}_i \in \mathcal{\tilde{B}}$, and $\mathbf{x}_j \in \mathcal{B}_r$}
            {
            \begin{minipage}[t]{0.43\textwidth}  %
            \tcc{ \hspace{10mm}  Local KD \hspace{5mm} }
             \begin{flushleft} \vspace{3mm}
             $\mathbf{z}_i, \mathbf{z}_j, \mathbf{\tilde{z}}_i \leftarrow f^{m}(\mathbf{x}_i), f^{m}(\mathbf{x}_j), f^{m}(\mathbf{\tilde{x}}_i)$ \\ \vspace{3.8mm} 
             \tcp{Local relationship vectors} \vspace{0.3mm}
             $\mathbf{r}_i^j = \frac{\exp(\text{sim}(\mathbf{z}_i, \mathbf{z}_j)/\tau)}{\sum_{k \in \mathcal{B}_r} \exp(\text{sim}(\mathbf{z}_i, \mathbf{z}_k)/\tau)}$ \vspace{2mm} \\ 
             $\mathbf{\tilde{r}}_i^j = \frac{\exp(\text{sim}(\mathbf{\tilde{z}}_i, \mathbf{z}_j)/\tau)}{\sum_{k \in \mathcal{B}_r} \exp(\text{sim}(\mathbf{\tilde{z}}_i, \mathbf{z}_k))/\tau)}$ \\ \vspace{0.5mm}
             $L_c^{\text{local}} \leftarrow$ Calculate from Eq.~\ref{eq:contrastive_loss} or~\ref{eq:byol_loss}\\
             $L_r^{\text{local}} \leftarrow$ Calculate from Eq.~\ref{eq:relational_loss} \\
             $L_{\text{local-KD}} = L_c^{\text{local}} + L_r^{\text{local}}$
            \end{flushleft}
            \end{minipage}
            \hspace{3mm}
            \begin{minipage}[t]{0.43\textwidth}  %
            \tcc{ \hspace{10mm}  Global KD \hspace{5mm} }
            \begin{flushleft}
            $\mathbf{z}_i^l, \mathbf{\tilde{z}}_i^l \leftarrow h^{m} \circ f^{m} (\mathbf{x}_i), h^{m} \circ f^{m} (\mathbf{\tilde{x}}_i)$ \\
            $\mathbf{z}_i^g, \mathbf{z}_j^g, \mathbf{\tilde{z}}_i^g \leftarrow F(\mathbf{x}_i), F(\mathbf{x}_j), F(\mathbf{\tilde{x}}_i)$ \vspace{2mm} \\  
            \tcp{Global relationship vectors}
            $\mathbf{r}_{i}'^j = \frac{\exp(\text{sim}(\mathbf{z}_i^l, \mathbf{z}_j^g)/\tau)}{\sum_{k \in \mathcal{B}_r} \exp(\text{sim}(\mathbf{z}_i^l, \mathbf{z}_k^g)/\tau)}$ \\
            $\mathbf{\tilde{r}}_{i}'^{j}  = \frac{\exp(\text{sim}(\mathbf{\tilde{z}}_i^l, \mathbf{z}_j^g)/\tau)}{\sum_{k \in \mathcal{B}_r} \exp(\text{sim}(\mathbf{\tilde{z}}_i^l, \mathbf{z}_k^g)/\tau)}$ \\
            $L_c^{\text{global}} \leftarrow$ Calculate from Eq.~\ref{eq:contrastive_loss_global} \\
            $L_r^{\text{global}} \leftarrow$ Calculate from Eq.~\ref{eq:global_relational_loss} \\
            $L_{\text{global-KD}} = L_c^{\text{global}} + L_r^{\text{global}}$ \\
            \end{flushleft}
            \end{minipage}
            }
            $L_{\text{total-KD}} = L_{\text{local-KD}} + L_{\text{global-KD}}$  \hfill \tcp{Calculate the total loss} 
            Update $f^m$, $h^m$ via back-propagation \hfill  \tcp{Update local model parameters}
        }
    }
    \textbf{return} $f^m$, $h^m$
\end{algorithm}

\subsection{Further Results on Performance Evaluation}
\subsubsection{Statistical errors in Table~\ref{tab:main_result}.}
Table~\ref{tab:main_result_with_ste} reports standard errors from various training iterations (i.e., last five epochs \& best five epochs), which shows that our model's accuracy converges stably.

\begin{table}[t!]
\caption{
Adding \model{} increases the performance of various federate learning models. The final round accuracy and the best accuracy are reported with standard errors.
}
\scalebox{0.93}{
\centering
{
\begin{tabular}{l|cc|cc|cc}
\toprule
\multirow{2}{*}{Method} & \multicolumn{2}{c|}{CIFAR-10} &  \multicolumn{2}{c|}{SVHN} & \multicolumn{2}{c}{F-MNIST} \\ 
                       & \hspace{2.2mm} Last \hspace{2.2mm}  & \hspace{2.2mm} Best \hspace{2.2mm}  & 
                       \hspace{2.2mm} Last \hspace{2.2mm} & \hspace{2.2mm} Best \hspace{2.2mm}  & \hspace{2.2mm} Last \hspace{2.2mm} & \hspace{2.2mm} Best \hspace{2.2mm} \\ \midrule
FedSimCLR             &   51.31$\pm$0.25     &  52.88$\pm$0.06      &   75.19$\pm$3.14    &  76.50$\pm$0.04       &  77.66$\pm$0.75      &    79.44$\pm$0.06   \\
+ \model{}           & \textbf{56.88$\pm$0.72} & \textbf{57.95$\pm$0.03} &   \textbf{77.19$\pm$1.67} & \textbf{77.70$\pm$0.09} &\textbf{81.98$\pm$0.74} & \textbf{82.47$\pm$0.02}   \\ \hline
FedMoCo               &	56.74$\pm$1.63 & 57.82$\pm$0.02 &	70.69$\pm$3.03 &  70.99$\pm$0.07 & 82.31$\pm$0.18 & 83.58$\pm$0.06 \\
+ \model{}            	&  \textbf{58.23$\pm$1.22} & \textbf{59.43$\pm$0.09}  & \textbf{73.57$\pm$2.79} & \textbf{73.92$\pm$0.02}  & \textbf{83.62$\pm$0.32}  & \textbf{84.65$\pm$0.04}  \\ \hline
FedBYOL              &	52.24$\pm$0.61  & 53.14$\pm$0.06	 	&65.95$\pm$1.62 & 67.32$\pm$0.24		&81.45$\pm$0.27 & 82.37$\pm$0.06 \\
+ \model{}           	&\textbf{56.49$\pm$0.72} & \textbf{57.79$\pm$0.12}& 	\textbf{68.94$\pm$1.13} & \textbf{69.05$\pm$0.07} &  \textbf{83.18$\pm$0.31} 	 & \textbf{84.30$\pm$0.07}   \\ \hline
FedProtoCL          	& 51.33$\pm$1.03 & 52.12$\pm$0.03 & 49.85$\pm$0.77 & 50.19$\pm$0.11	&  81.76$\pm$0.22 & \textbf{83.57$\pm$0.04}   \\
+ \model{}             &  \textbf{55.36$\pm$0.98}   & \textbf{56.76$\pm$0.01}   &   \textbf{69.31$\pm$1.72}    &   \textbf{69.75$\pm$0.15}   &    \textbf{82.74$\pm$0.35}    &   83.34$\pm$0.04    \\ \hline
FedU                   &  50.79$\pm$0.47   & 50.79$\pm$0.05         &   66.02$\pm$1.83  &  66.22$\pm$0.17     &  80.59$\pm$0.42     &   82.03$\pm$0.05         \\
+ \model{}                &  \textbf{56.15$\pm$0.58}   &     \textbf{57.26$\pm$0.05}  &     \textbf{68.13$\pm$1.17}    &  \textbf{68.39$\pm$0.07}   &   \textbf{83.73$\pm$0.20}    &   \textbf{84.12$\pm$0.03}    \\ \bottomrule
\end{tabular}}
}
\label{tab:main_result_with_ste}
\end{table}

\begin{table}[t!]
\caption{Performance improvement with three different algorithms on classification accuracy. Both the final round accuracy and the best accuracy show that \model{} brings nontrivial improvement over the baseline algorithm.}
\centering{
\scalebox{1}{
\begin{tabular}{l|cccccc}
\toprule
\multirow{2}{*}{Method} & \multicolumn{2}{c}{CIFAR-10} & \multicolumn{2}{c}{SVHN} & \multicolumn{2}{c}{F-MNIST} \\
                        & ~~Last~~        &   ~~Best~~          &  ~~Last~~    & ~~Best~~     &  ~~Last~~      &  ~~Best~~       \\ \midrule
FedSimCLR  & 51.31 & 52.88 & 75.19 & 76.50 & 77.66 & 79.44 \\ 
 + FedCA &  47.46 & 48.54  & 59.40 & 59.86 &  81.51 & 82.05 \\ 
 + MOON-unsup & 51.78   & 52.84  & 75.36 & 76.03 & 80.58 & 80.93 \\  \midrule
 + \model{} (ours) &   \textbf{56.88} & \textbf{57.95}    & \textbf{77.19} & \textbf{77.70} &  \textbf{81.98} & \textbf{82.47}\\\bottomrule
\end{tabular}}}
\label{Tab:classification}
\end{table}

\begin{table}[!t]
\centering
\caption{Performance improvement on ImageNet-10. Both last and best round accuracy are reported.}
\begin{tabular}{l|cccc}
\toprule
\multirow{2}{*}{Method} & \multicolumn{2}{c}{FedSimCLR} &  \multicolumn{2}{c}{FedSimCLR+\model{}} \\
                          &  Last        &   Best &   Last    & Best  \\\midrule
 ImageNet-10 ~~ &  ~~ 81.50  ~~ &  ~~ 81.50  ~~ &   ~~ \textbf{86.17}  ~~ &  ~~ \textbf{86.57}  ~~ \\
\bottomrule 
\end{tabular}
\label{Tab:ImageNet}
\end{table}

\subsubsection{Comparison with other relevant baselines.}
We compared \model{} with other contrastive learning methods: FedCA (Zhang \textit{et al.})\footnote{Note that the dictionary module has been removed from FedCA for fair comparison as it directly shares the local data information of all clients.} and MOON (Li \textit{et al.}). We followed the same implementation guidelines as in the original work, only substituting the InfoNCE objective for the supervised loss in MOON (calling this MOON-unsup). FedSimCLR with FedAvg served as the base framework in both cases. Our method continues to outperform, as shown in Table~\ref{Tab:classification}, demonstrating the benefit of the proposed knowledge distillation strategy in unsupervised federated learning.

\subsubsection{Performance on ImageNet.}
To verify our model's applicability to the large-scale dataset, we run \model{} on ImageNet-10 benchmark, a 10-class subset of ImageNet. \model{} was trained for 10 local epochs in each communication round, with a total of 50 rounds. Table~\ref{Tab:ImageNet} shows that adding \model{} to FedSimCLR improves the classification accuracy by 5pp. Extended results for the larger number of the class will also be released.


\subsubsection{Full comparison results over communication rounds.}
\begin{figure}[t!]
\centering
\begin{subfigure}[t]{0.96\textwidth}
      \centering\includegraphics[width=0.99\textwidth]{figures/fedsimclr_round.png}
      \subcaption{Performance gain on FedSimCLR}
      \label{fig:further_fedsimclr_round}
\end{subfigure}
\begin{subfigure}[t]{0.96\textwidth}
     \centering\includegraphics[width=0.99\textwidth]{figures/fedbyol_round.png}
      \subcaption{Performance gain on FedBYOL}
      \label{fig:further_fedbyol_round}
\end{subfigure}
\begin{subfigure}[t]{0.96\textwidth}
        \centering\includegraphics[width=0.99\textwidth]{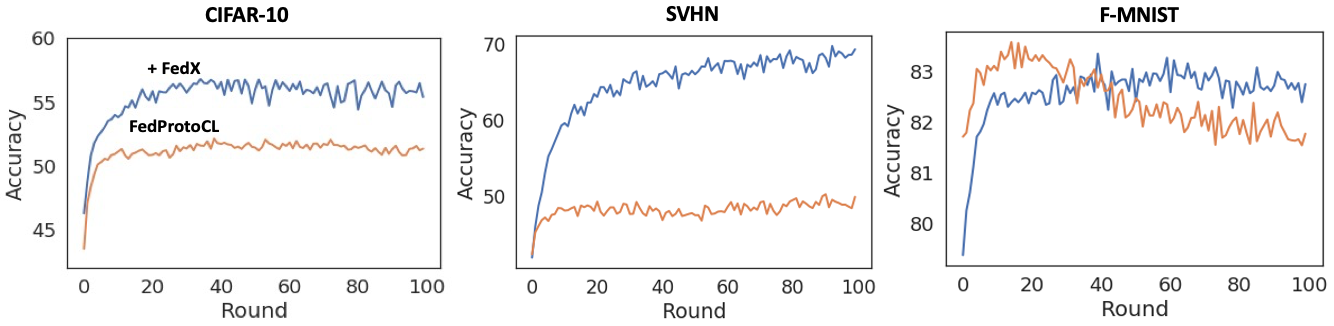}
      \subcaption{Performance gain on FedProtoCL}
      \label{fig:further_fedprotocl_round}
\end{subfigure}
\begin{subfigure}[t]{0.96\textwidth}
     \centering\includegraphics[width=0.99\textwidth]{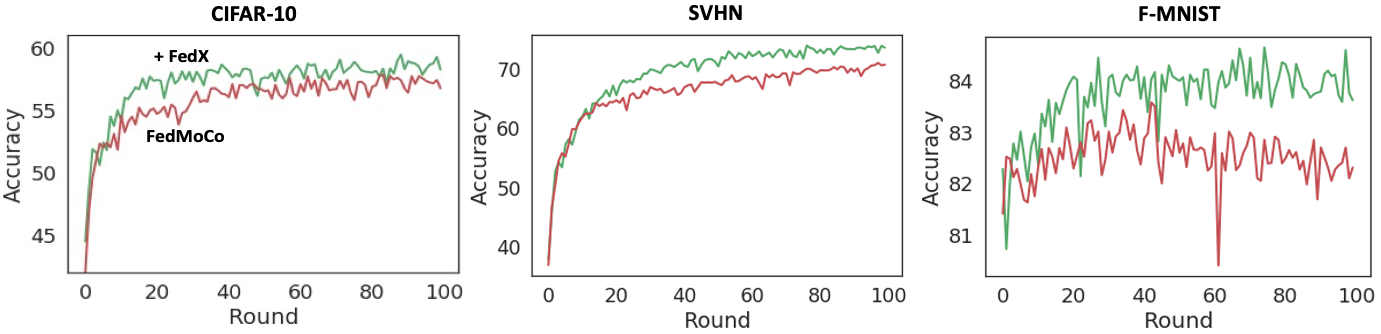}
      \subcaption{Performance gain on FedMoCo}
      \label{fig:further_fedmoco_round}
\end{subfigure}
\begin{subfigure}[t]{0.96\textwidth}
        \centering\includegraphics[width=0.99\textwidth]{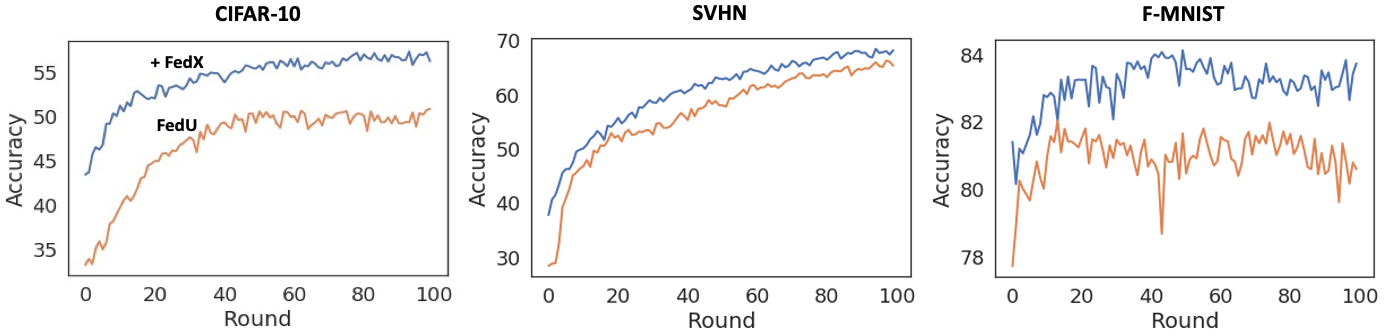}
      \subcaption{Performance gain on FedU}
      \label{fig:further_fedu_round}
\end{subfigure}
\caption{Performance comparison between all baselines (i.e., FedSimCLR, FedBYOL, FedProtoCL, FedMoCo, and FedU) and \model{}-enhanced versions over communication rounds. \model{} brings extra performance on baseline unsupervised learning models in three benchmark datasets.} 
\label{fig:further_evaluation}
\end{figure}

\model{} brings meaningful performance improvements than when using the baseline algorithms alone. Figure~\ref{fig:further_evaluation} shows trajectories including three additional baselines -- FedProtoCL, FedMoCo, and FedU that were omitted in Fig.~\ref{fig:round}. We can check how quickly the model benefits baselines over the varying communication rounds. Especially, in FedSimCLR and FedProtoCL (F-MNIST) experiments, \model{} prevents the local bias degrading the performance during the entire training phase and thereby stops such deterioration in performance.

\end{document}